\documentclass[letterpaper, 10 pt, conference]{ieeeconf}  
\IEEEoverridecommandlockouts       
\usepackage[english]{babel}

\usepackage{graphicx}
\usepackage{animate}
\usepackage{epsfig} 				
\usepackage{epstopdf}

\usepackage{listings}
\usepackage{color}
\usepackage{nameref}
\usepackage{hyperref}
\usepackage{amsmath}	 			
\usepackage{amssymb}  				

\usepackage{dsfont}			
\usepackage{mathtools}

\usepackage{epigraph}
\usepackage{lscape}
\usepackage[]{nomencl}				
\usepackage{algorithm}
\usepackage{algorithmic}
\usepackage{multicol}
\usepackage{multirow}
\usepackage{etoolbox}

\usepackage{caption}
\usepackage{subcaption}
\usepackage{wrapfig}

\usepackage{siunitx}

\usepackage{floatflt}

\usepackage{url}

\let\labelindent\relax
\usepackage{enumitem}
\usepackage{booktabs}
\usepackage{cleveref}

\newcommand{\email}[1]{\href{mailto:#1}{\nolinkurl{#1}}}

\newcommand{\link}[1]{\colora{\url{#1}}}

\renewcommand{\sec}[1]{Section~\ref{#1}}
\newcommand{\fig}[1]{Fig.~\ref{#1}}
\newcommand{\eq}[1]{Equation~\eqref{#1}}

\newcommand{\tab}[1]{Table~\ref{#1}}

\newcommand{\titlelong}[0]{Towards Learning to Play Piano with Dexterous Hands and Touch}

\newcommand{\E}{\mathbb{E}}

\usepackage{bm}
\newcommand{\citet}[1]{\cite{#1}}
\newcommand{\citep}[1]{\cite{#1}}
\title{\LARGE \bf \titlelong{}}
\author{Huazhe Xu$^{1,*}$, Yuping Luo$^{2}$, Shaoxiong Wang$^{3}$, Trevor Darrell$^{4}$, Roberto Calandra$^{5}$
\thanks{$^{*}$ Work done while at UC Berkeley and Meta AI.}
\thanks{$^{1}$ IIIS, Tsinghua University {\tt\small xuhuazhe12@gmail.com}, $^{2}$ Princeton University, $^{3}$ MIT, $^{4}$ UC Berkeley, $^{5}$ Meta AI\newline}
}

\begin{document}

\maketitle
\thispagestyle{empty}
\pagestyle{empty}

\begin{abstract}
	As Liszt once said ``(a virtuoso) must call up scent and blossom, and breathe the breath of life'', a virtuoso plays the piano with passion, poetry, and extraordinary technical ability.
Hence, piano playing, being a task that is quintessentially human, becomes a hallmark for roboticians and artificial intelligence researchers to pursue.
In this paper, we advocate an end-to-end reinforcement learning~(RL) paradigm to demonstrate how an agent can learn directly from machine-readable music score to play the piano with touch-augmented dexterous hands on a simulated piano.
To achieve the desired tasks, we design useful touch- and audio-based reward functions and a series of tasks. 
Empirical results show that the RL agent can not only find the correct key position but also deal with the various rhythmic, volume, and fingering requirements.
As a result, the agent demonstrates its effectiveness in playing simple pieces that have different musical requirements which show the potential of leveraging reinforcement learning approach for the piano playing tasks. 

\end{abstract}



\section{INTRODUCTION}

	Piano playing is a complex sensorimotor task that requires a combination of hitting the correct keys with certain amount of force, and accurate timing. 
It takes pianists years and decades to master the skills; hence, it draws a critical line between everyday tasks such as objects grasping and piano playing. 
Enabling robot hands to play the piano is a challenging and meaningful task that might involve multiple lines of research.
Reinforcement Learning~(RL) has been proved to outperform humans in games such as Atari~\citep{mnih2013playing}, Go~\citep{silver2018general} and achieve impressive results on continuous control tasks~\citep{andrychowicz2020learning}. 
However, RL has seldomly been applied on instrument playing tasks. 
This can be attributed to three aspects: 1) The lack of simulation environments for instrument playing. 2) The lack of multi-modal sensory inputs such as touch signals, which play an important role in many instruments playing tasks. 3) The challenging nature of the tasks that involves rhythm, pitch, musicality, etc. 
In this paper, we explore how RL agents can play the piano with the help of image-based tactile sensory input.

 There is a long history of attacking the automatic piano playing task. Previous works~\citep{pianola, dexterousfinger, railpiano, railpiano2, topper2019piano} relied on specific design of robot hand or specialized mechanical setups. Another branch of works~\citep{controllerdesign, li2014intelligent, li2017adaptive, scholz2019playing} tried to solve the task with commercial robot hands. However, most of them were manually programmed and hence lack of the ability to adapt and generalize to new pieces. There is also work~\citep{Hugheseaau3098} that used a human-like hand to play the piano; however, the fingers are passive and unable to play a piece.

\begin{figure}[t]

  \centering
  \includegraphics[width=\columnwidth]{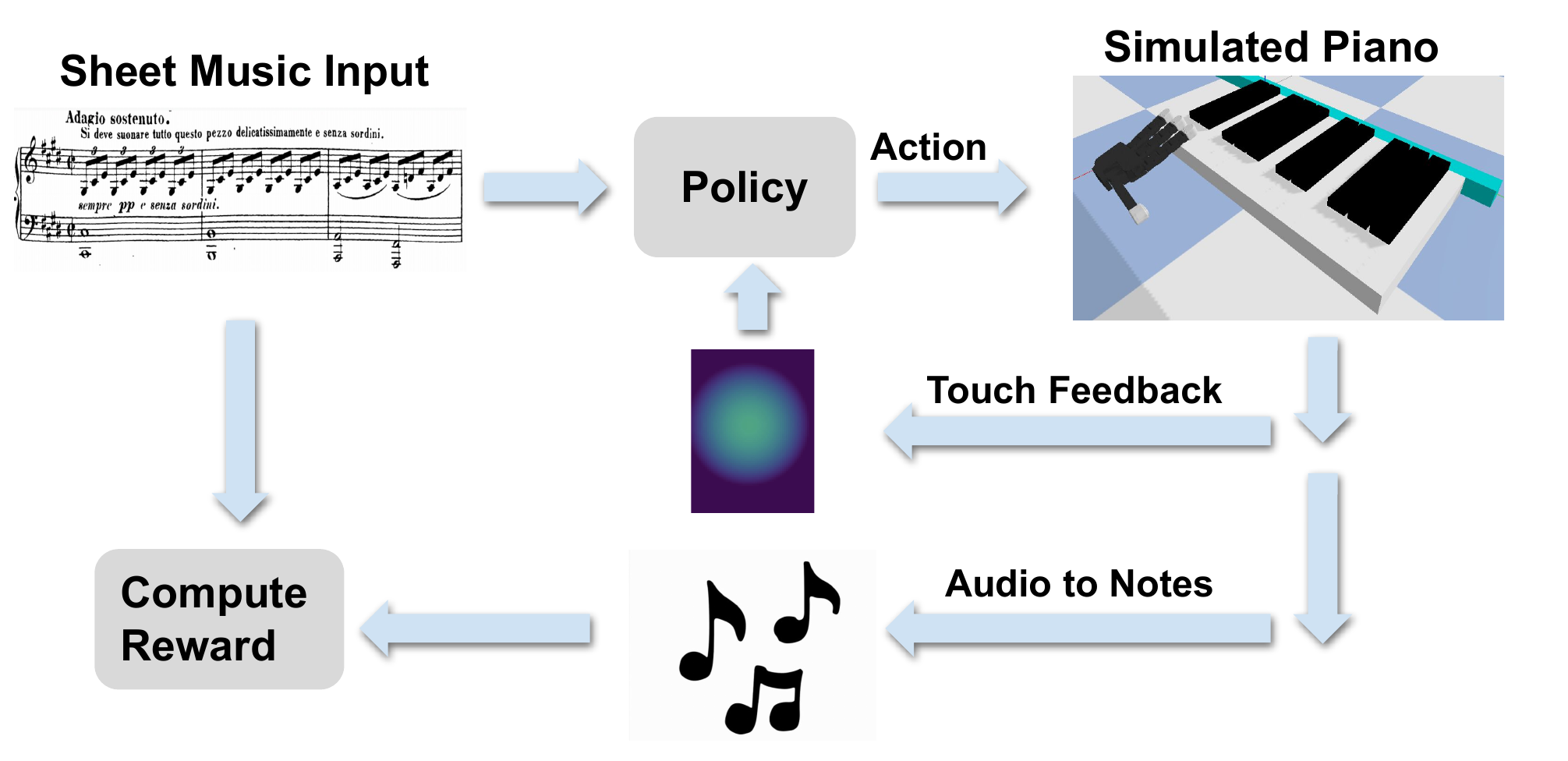}
  \caption{Playing the piano is intrinsically a multi-modal task involving vision, audio and touch. Using reinforcement learning, we demonstrate that it is possible for an agent to learn to play the piano by exploiting tactile sensors for control and using the corresponding notes and touch feedback for computing rewards.}
  \label{fig:teaser}
\label{fig:fig}
\end{figure}

In this work, we study how to leverage reinforcement learning algorithms and tactile sensor to tackle dexterous hands piano playing tasks, in a simulated environment, as sketched in \fig{fig:teaser}. We first formulate the piano playing task as a Markov Decision Process~(MDP) so that we can apply reinforcement learning algorithms. We also explore the possibilities of different reward function designs which is usually hard to choose to empower a successful RL algorithm. More specifically, we train an end-to-end policy to directly output actions for controlling the hand joints and accomplishing what is presented in the music score. We further study how to integrate rich tactile sensory input to the proposed method as well as to what extent tactile data help with the task.

Our contribution in this work can be summarized as follows:
1) Based on the TACTO touch simulator~\citep{wang2020tacto}, we build a hand-piano simulator which includes a multi-finger allegro hand equipped with a simulated DIGIT tactile sensors, and a small size piano keyboard. The piano automatically translate the physical movement of keys into MIDI signals.
2) We are the first to formulate the piano playing task as an MDP and use reinforcement learning algorithms on simulated piano playing tasks. This opens a welcoming avenue for benchmarking state-of-the-art RL algorithms on complex and artistic tasks.
3) We propose to use high-resolution vision-based tactile sensors to improve piano playing especially on fingering indication, which provide useful knowledge on how much tactile data would help on such tasks and thus be adopted as a common paradigm on related future tasks.


\section{RELATED WORK}
\label{sec:related}
	Piano playing has been studied for decades as both a real-world task for robot hand control and a hallmark toward more general applications.  Starting from the early player pianos called Pianola~\cite{pianola} that requires pumping pedal with recorded score on paper rolls, researchers have developed robotics systems that enable a robot hand and infer the required action based on the input signals~\cite{dexterousfinger,controllerdesign,inputscore}. To enhance the robustness of such systems, various efforts has been made such as adding a rail to control the degree of freedom~\cite{railpiano,railpiano2}, reducing the number of fingers~\cite{minihand} for easier control or even adding more fingers~\cite{eightfingers} on top of each of the keys. Some works also explore the use of a human-like hand to play the piano; however, the hand is either passive without force control on the fingers~\cite{Hugheseaau3098} or omit the velocity~\cite{TopperMaloney2019}. The reverse problem is also well-studied in the previous work that tracks piano performance based on human players~\cite{reverse}. In contrast to most of the previous work, our method uses a dexterous hand that is \textit{not} specialized for piano playing to enable a \textit{learning} agent to play the piano with the help of multi-modal sensory signals. When evaluating the performance, we also take the resulting key velocity into consideration for artistic playing where only limited work~\cite{forcecontrol} also take care of such an aspect.

It is also the first time reinforcement learning is used to enable a robot hand to play the piano without manually designed motion planning algorithms.
In previous work~\cite{rajeswaran2017learning, andrychowicz2020learning}, RL has been used to accomplish manipulation tasks such as open door or object manipulation. The recent work~\cite{akkaya2019solving} has also demonstrate simulator trained RL policies can be transfered to real robot hands on rubik's cube manipulation. 
While RL has been successful applied on many robotics hand control tasks, most of them only require the agent to perform correct joint positions.
However, our work is trying to design effective RL algorithm to achieve not only correct joint positions but also temporal correctness and force in instrument playing tasks. Such tasks are drastically different from the previous ones and can be used as service robot or delegate for harder dexterous hand control tasks.

Vision-based tactile sensors~\cite{yuan2017gelsight, Lambeta2020DIGIT} has been developed to provide high-resolution touch signals with high frame rate. In previous works, they demonstrated their potential to improve the quality for robot hands in sensorimotor tasks~\cite{calandra2018more,Lambeta2020DIGIT}. Recent advances also developed robot systems that can control hands to perform challenging tasks such as manipulating cables~\cite{she2019cable} or swinging a bottle~\cite{wang2021swingbot}. However, in this paper, we rely on the tactile sensor for recognizing which finger is being used. We also find augmenting the whole system with tactile sensors would make the algorithm converge fast due to the additional information. Thanks to the opensourced tactile sensor TACTO~\cite{wang2020tacto}, we can incorporate virtual sensors with the robot hand for better piano performance.


\section{REINFORCEMENT LEARNING OF PIANO PLAYING}
\label{sec:approach}

\begin{figure*}[t]

  \centering
  \includegraphics[width=0.98\linewidth]{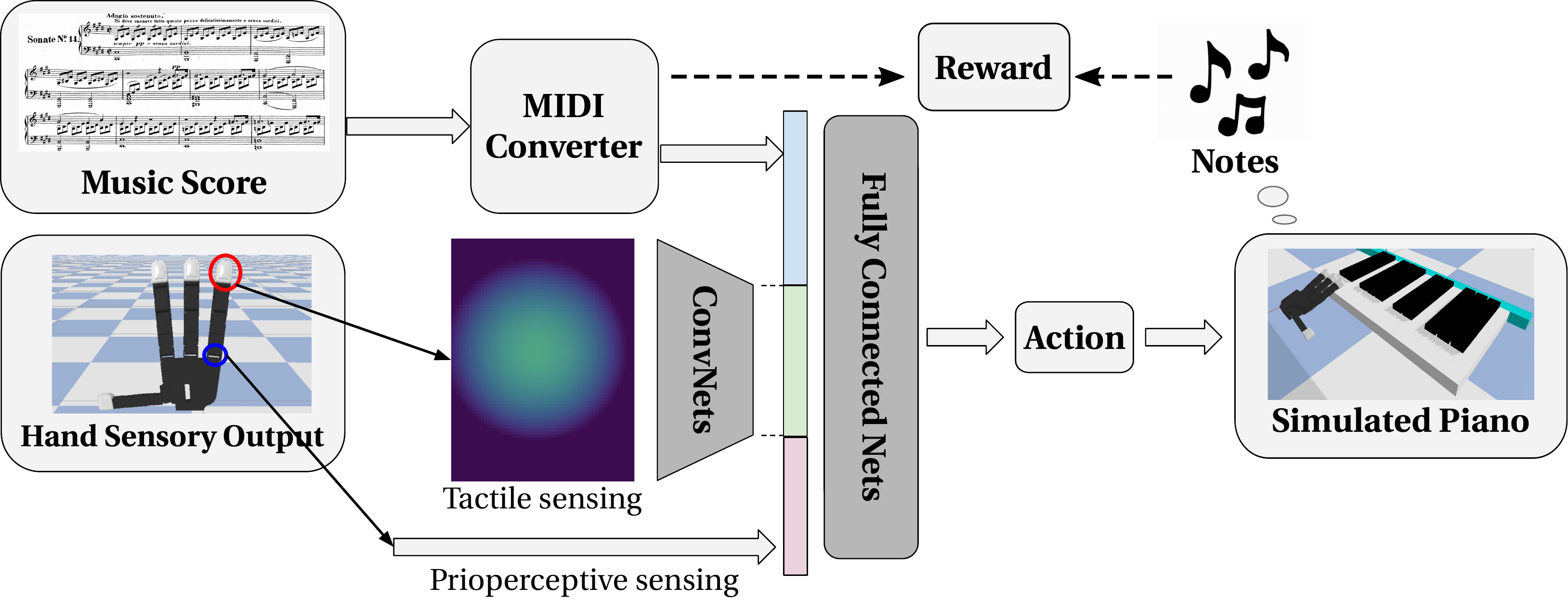}
  \caption{The overview of the proposed method. The observation space is composed of three components: 1) vectorized MIDI sheet music which is converted through a MIDI converter, 2) tactile sensory data from DIGIT sensor processed by convolutional neural network, 3) robot hand joint and wrist information extracted from the simulator. We use a fully connected neural net as the policy network that interacts with the simulated piano. Then we compute the reward function based on the MIDI music and the output audio from the piano.}
  \label{fig:sub-first}

\label{fig:overview}
\end{figure*}

	In this section, we introduce our system for enabling the agent to play the piano. In \fig{fig:overview}, we use a schematic diagram to demonstrate the system. In \sec{sec:formulation}, we first formulate the task as a Markov decision process and detail about the components of the MDP. In \sec{sec:score}, we introduce the representation for music score. In \sec{sec:sac}, we introduce the off-policy RL algorithm we used for this task, the neural network architecture, and training procedure. 

\subsection{Formulation}
\label{sec:formulation}
We formalize piano playing as a Markov decision process~(MDP)~\citep{Bellman1957Markovian} where we select the actions that minimize the difference between the attacked key and target key from a digital music score. In this MDP, at every time step $t$, a robot has the observation $\boldsymbol{o}_t$ from the observation space~$\mathcal{O}$. The agent is supposed to choose an action $\boldsymbol{a}_t$ from the action space~$\mathcal{A}$ that directs each joint to a new pose relative to its current pose. The unknown transition of the environment~(piano) can be expressed with probability density of the next observation  $\boldsymbol{p}: \mathcal{O}\times\mathcal{A}\times\mathcal{O} \rightarrow [0, \infty)$. The environment emits a bounded reward $r: \mathcal{O}\times\mathcal{A}\rightarrow[r_\text{min}, r_\text{max}]$. The reward is computed based on human priors about how to evaluate the played keys when the target music score is available.

Under this formulation, we describe the observation space, action space, reward structure below:
\subsubsection{Observation Space} \label{sec:obs} The robot's observation space includes the MIDI score, recent history of the image-based tactile sensory outputs from each finger, robot hand positions, robot finger joint angles, velocities, and piano key angles: $\boldsymbol{o}_t = \{\boldsymbol{o}_t^{MIDI},\boldsymbol{o}_{t-1}^{\textrm{tacto}}, \boldsymbol{o}_t^{\textrm{tacto}},\boldsymbol{o}_t^{\textrm{hand\_position}}, \boldsymbol{o}_t^{\textrm{hand\_joint\_angles}}, \boldsymbol{o}_t^{\textrm{hand\_joint\_vel}}, \allowbreak \boldsymbol{o}_t^{\textrm{key\_joint\_angles}}, \boldsymbol{o}_t^{\textrm{key\_vel}} \}$. Although more information might give better performance, in practice, we exclude all the key information except for ablation study.  

\noindent \textbf{MIDI input.} The robot observes the the MIDI input at the current timestep. The notes are one-hot vectorized. 

\noindent \textbf{Tactile image.} The robot observes the history of tactile sensory outputs over the last 2 timesteps, which are $60 \times 80$ depth-like images $\boldsymbol{o}_t^{\textrm{tacto}} \in \mathbb{R}^{60 \times 80}$.

\noindent \textbf{Hand position.} The robot observes the Cartesian positions of the wrist of the hand. We fix the hand height, so it becomes two dimensional observation   $\boldsymbol{o}_t^{\textrm{hand\_position}} \in \mathbb{R}^2$.

\noindent \textbf{Hand joint angles and velocities.} The robot observes the current hand joint angles and optionally velocities. In our setup, there are in total $12$ joints in use for the allegro hand: $\boldsymbol{o}_t^{\textrm{hand\_joint\_angles}} \in \mathbb{R}^{12}$,  $\boldsymbol{o}_t^{\textrm{hand\_joint\_vel}} \in \mathbb{R}^{12}$.

\noindent \textbf{Key joint angles and velocities.} The robot optionally observes the current piano key joint angles and velocities. We note that these are privileged knowledge in real world piano. We add it mainly for ablation purpose. $\boldsymbol{o}_t^{\textrm{key\_joint\_angles}} \in \mathbb{R}^{12}$, $o_t^{\textrm{key\_vel}} \in \mathbb{R}^{12}$. 

\subsubsection{Action Space} \label{sec:action} Given the observations elaborated in \sec{sec:obs}, the robot learns a parameterized policy $\pi$ to generate actions comprising of 1) movement of fingers joints, 2) movement of hand position. $\boldsymbol{a}_t = \{\boldsymbol{a}_t^{\textrm{finger\_1}}, \boldsymbol{a}_t^{\textrm{finger\_2}}, \boldsymbol{a}_t^{\textrm{finger\_3}}, \boldsymbol{a}_t^{\textrm{wrist}}\}$.

\noindent \textbf{Finger movements.} The robot controls each joint of the fingers with either a specified velocity or a torque: $\boldsymbol{a}_t^{\textrm{finger}_i} \in [-1, 1]$ where $-1$ indicates to lift the finger with maximum speed/torque, and $1$ means put down the finger with maximum speed/torque. 

\noindent \textbf{Hand movement.} The robot moves the hand wrist position. Hence, the hand movement action $\boldsymbol{a}_t^{\textrm{wrist}} \in [-1, 1]$. $-1$ indicates moving toward the left and $1$ indicates moving toward the right. To alleviate exploration issues, we also discretize the action space. 

\subsubsection{Reward Structure} \label{sec:reward} One important building block for using RL agents to play the piano is an informative reward function. An ideal reward function should include a few things: 1) It penalizes when a key is not attacked when the score indicates it should be.  (2) It penalizes when the correct key is hit but with a wrong velocity. We first introduce the most general reward as
\begin{align}
\label{eq:gen_rew}
r(\boldsymbol{o}, \boldsymbol{a}) = \textrm{Const} -\sum_i^{\mid K \mid} d(\tilde{k}_i, k_i)\,,
\end{align}
where $K$ is the set of all the keys that is possible to be attacked. $| K |$ indicates the cardinality of set $K$.   $d(\cdot, \cdot)$ is a distance metric that measures the difference between normalized velocities of two keys
\begin{align}
d(\widetilde{k_i}, k_i) = |v_{\widetilde{k_i}} - v_{{k}_i}|\,.
\end{align}
We add a small constant so that even when the agent plays one incorrect key, the reward is still positive. This is important for encouraging the agent to play a key rather than doing nothing. We also note that the distance function can be chosen based on empirical performance.

We then introduce a more informative dense reward function to guide the agents for harder tasks. We use $S$, $X$, $T$ to represent the played keys, imagined keys and target keys (in the music score) at every time step instead of using $K$ for all the possible keys.
We want to design a distance function between $S$ and $T$ such that for any $T$, the minimizer of such loss w.r.t. $S$ is unique and equals to $S = T$. 
Hence, we will have the dense reward function
\begin{align}
\label{eq:w_rew}
r(\boldsymbol{s}, \boldsymbol{a}) = \text{Const} -W_{d, \sigma}(S, T)\,,
\end{align}
where the distance function $W_{d,\sigma}$ can be written as
\begin{equation}
  W_{d, \sigma}(S, T) = \min_{|X| = |T|} \left[ \sum_i d(X_i, T_i) + \sigma |X \triangle S| \right]\,.
  \label{eq:wasserstein}
\end{equation}

 Our distance proposal here is inspired by the \emph{earth mover's distance}~\citep{vallender1974calculation}. Suppose we want to modify $S$ to $T$ with three possible operations: a) remove an item in $S$ with cost $\sigma > 0$; b) add an item in $S$ with cost $\sigma$; c) modify an item in $S$ to an item in $T$ with cost $d(\cdot, \cdot)$.
The distance between $S$ and $T$ can then be defined by the minimal cost among all modification scheme from $S$ to $T$.
We note that all the three operations have corresponding operation in the real world: a) release a piano key, b) attack a piano key, and c) move from one attacked key to another.
The optimal modification scheme is then given by \eq{eq:wasserstein}.
More precisely, we break the optimal modification scheme into two steps: first we add or remove items, then we move items.
The imagined keys $X$ in \eq{eq:wasserstein} is essentially the modified $S$ after adding or removing keys.
The cost from $S$ to $X$ is $\sigma |X \triangle S|$, and the cost from $X$ to $T$ is $\sum_i d(X_i, T_i)$.
The optimization w.r.t. $X$ is done using SciPy~\citep{2020SciPy-NMeth}.

\subsection{Music Score Representation}
\label{sec:score}
Music can be represented by waveform or spectrogram. However, it can be hard for a robot to take continuous input and to design reward functions with these representations. Hence, we choose the Musical Instrument Digital Interface~(MIDI) as the audio representation. MIDI has timing information for note-on and note-off events and velocity information for each activated note. We freeze the tempo information to $60$ beats per minute. To augment the MIDI representation, we also include fingering information that indicates which finger to use for a note. We also note that fingering indicators are usually provided even for humans when it is hard to explore and find the optimal solution. 

 In practice, for each timestep, we use the matrix $M \in \mathbb{R}^{L \times 5}$ to represent the score. Here $L$ represents number of keys and $5$ dimensions for note (first dimension), velocity (second dimension), and fingering information (3-dim one-hot vector). For the note dimension, we use binary variable in $\{1, 0\}$ to indicate which note to play. In the velocity dimension, we use a float number in range $[0, 1]$. In the fingering dimension, we use an one-hot vector.


\subsection{Soft Actor-Critic}
\label{sec:sac}
\subsubsection{Algorithm}
We use reinforcement learning algorithm Soft Actor Critic~(SAC)~\citep{sac} for training the policy. It is an off-policy RL method using the actor-critic framework.
There are three types of parameters to learn in SAC: i) the policy parameters ${\phi}$; ii) a temperature $\alpha$; iii) two sets of the soft $Q$-function parameters $\theta_1, \theta_2$ to mitigate overestimation \citep{fujimoto2018addressing}. We can represent the policy optimization objective as
\begin{align}
J_\pi(\phi) = \E_{\boldsymbol{o}_t\sim\mathcal{D}}\left[\E_{\boldsymbol{a}_t\sim\pi_{\phi}}[\alpha \log \pi_{\phi}(\boldsymbol{a}_t|\boldsymbol{s}_t) - Q_{\theta}(\boldsymbol{o}_t, \boldsymbol{a}_t)]\right]\,
\label{eq:reparam_objective}
\end{align}
where $\alpha$ is a learnable temperature coefficient, and $D$ is the replay buffer. It can be learned to maintain the entropy level of the policy
\begin{align}
J(\alpha)  = \E_{\boldsymbol{a}_t \sim \pi_{\phi}} \left[ - \alpha\log\pi_{\phi}(\boldsymbol{a}_t|\boldsymbol{o}_t) - \alpha \bar{\mathcal{H}}\right],
\label{eq:ecsac:alpha_objective}
\end{align}
where $\bar{\mathcal{H}}$ is a desired entropy. The soft $Q$-function parameters $\theta_i$, $i \in \{1, 2\}$, can be trained by minimizing the soft Bellman residual as,
\begin{align}
J_Q(\theta_i) = \E_{(\boldsymbol{o}_t,\boldsymbol{a}_t)\sim\mathcal{D}} \left[ \frac{1}{2} (Q_{\theta_i}(\boldsymbol{o}_t,\boldsymbol{a}_t)-\hat{Q}(\boldsymbol{s}_t,\boldsymbol{a}_t))^2  \right],
\label{eq:Qfunc}
\end{align}
where
\begin{align*}
\hat{Q}(\boldsymbol{o}_t,\boldsymbol{a}_t) & = r(\boldsymbol{o}_t,\boldsymbol{a}_t) +\gamma \mathbb{E}_{\boldsymbol{a}_{t+1} \sim \pi_\phi(\boldsymbol{o}_{t+1})}\Big[  \\
  & \quad \min_{j \in \{1, 2\}} Q_{\bar\theta_j}(\boldsymbol{o}_{t+1},\boldsymbol{a}_{t+1}) - \alpha \log \pi_\phi(\boldsymbol{a}_{t+1} | \boldsymbol{s}_{t+1})\Big],
\label{eq:Qfunc2}
\end{align*}
and where $Q_{\bar\theta_j}$'s are the target networks \citep{lillicrap2015continuous}. The policy network is a neural network that computes $\boldsymbol{a} = f_\theta(\boldsymbol{o})$. .

\subsubsection{Architecture}
The policy network has three convolutional layers for image inputs and three multi-layer perceptron~(MLP) layers for vectorized state inputs. We use ReLU as the activation function. The features are then concatenated and fed into another two-layer MLP network that outputs the actions. We tried various design choices such as batch normalization layers or substituting the convolutional neural network with bilinear image downsampler. We find these design choices have minor influences on performance. We attribute this to the hard exploration challenge underlying in this task. 

\subsubsection{Training}
We use adam optimizer~\citep{kingma2014adam} with a learning rate $1e-3$. We initially set the tunable temperature variable $\alpha=0.02$. We allow $15000$ exploration steps before training starts for better exploration. 


\section{EXPERIMENTAL RESULTS}
\label{sec:result}

	In this section, we evaluate the RL-based agents on various meaningful piano playing tasks. We also ablate the model to recognize core factors for training such agents. 
In particular, we try to answer the following questions:

\begin{itemize}[noitemsep,topsep=0pt,parsep=0pt,partopsep=0pt,leftmargin=*]
    \item Can we use RL to learn to play the piano and produce correct notes, rhythm, chords and fingering?
    \item Can we handle the long-horizon nature of the task? 
    \item What are the important considerations for training such agents i.e., design choices, useful states? 
\end{itemize}

\subsection{Simulation}
\begin{figure}[t]
\begin{subfigure}{.49\columnwidth}
  \centering
  \includegraphics[height=2.4cm]{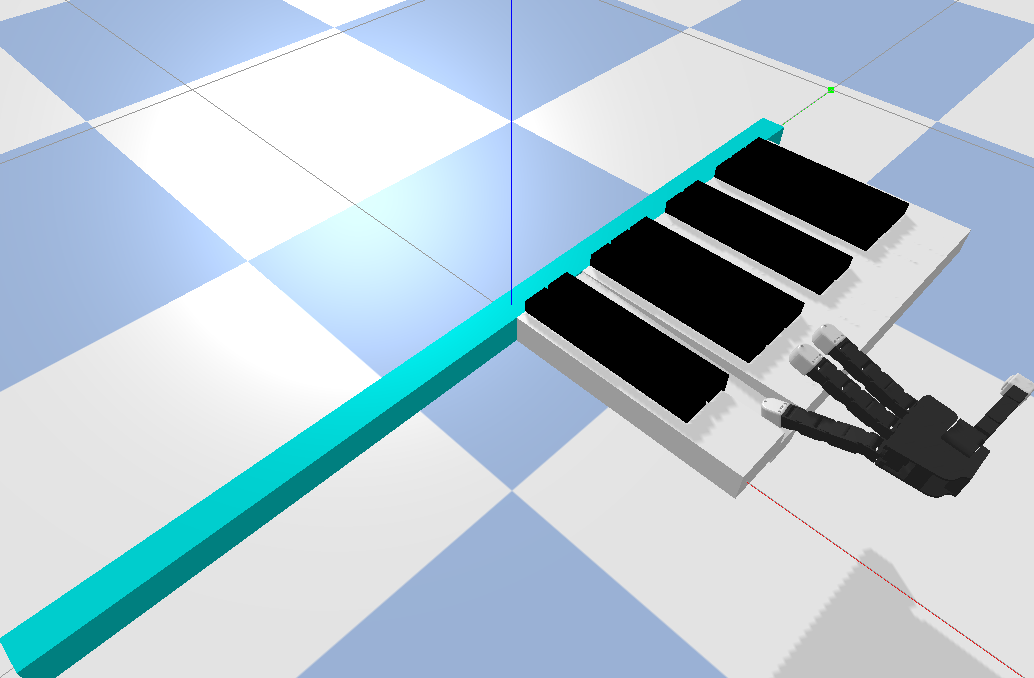}
  \label{fig:demo1}
\end{subfigure}
\hfill
\begin{subfigure}{.49\columnwidth}
  \centering
  \includegraphics[height=2.4cm]{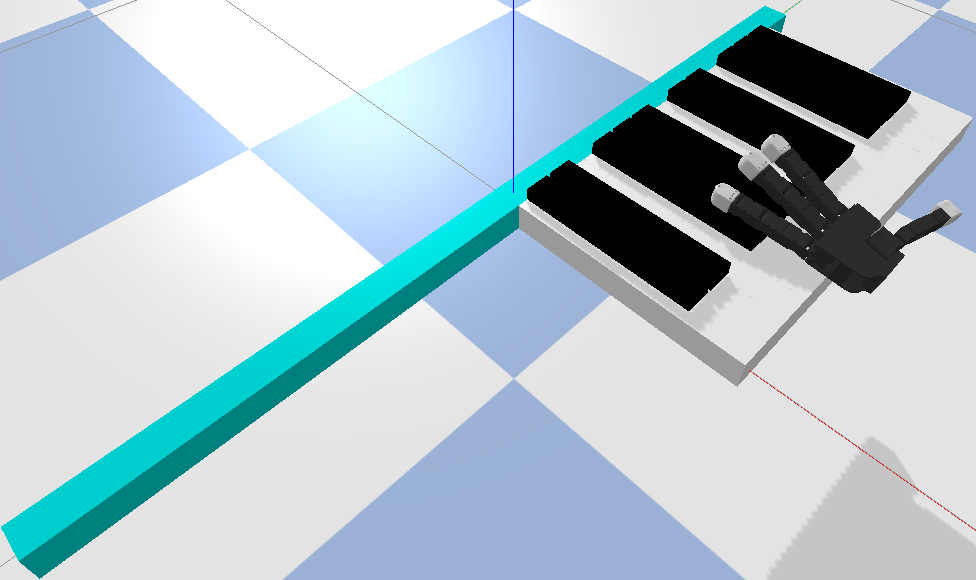}
  \label{fig:demo2}
\end{subfigure}

\caption{Samples from the piano-robot hand simulator. (\textit{Left}) Hand playing white keys of the piano. (\textit{Right}) Hand playing black keys of the piano.}
\label{fig:demo}
\end{figure}

\begin{figure*}[ht!]
\centering
\includegraphics[width=\linewidth]{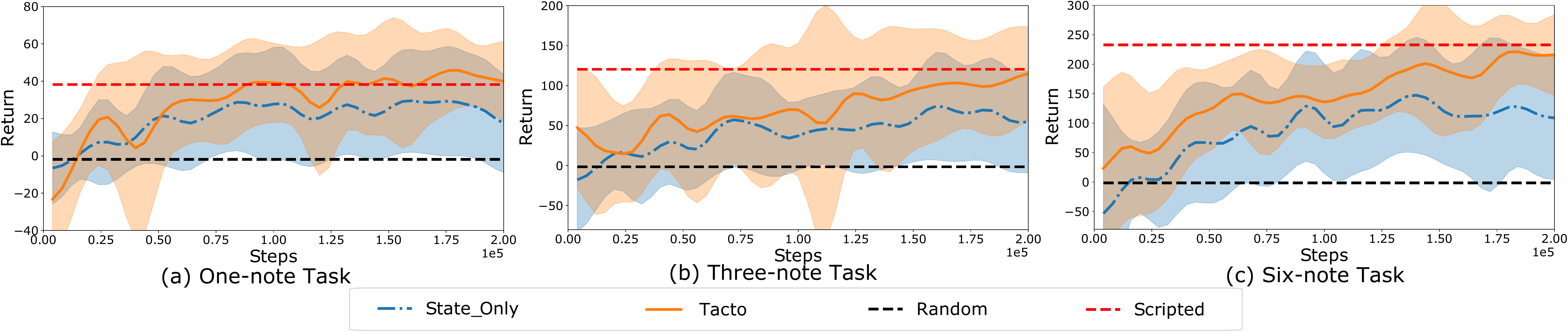}
\caption{Comparison on different levels of music tasks. All the experiments are run with 5 seeds. Shade is the one standard deviation. We find that RL-based algorithms can match the performance of manually designed controller. Tactile inputs improves the performance slightly.}
\label{fig:Q1_1}
\end{figure*}

\begin{figure}[ht!]
\begin{subfigure}{0.48\columnwidth}
  \centering
  \includegraphics[width=\linewidth]{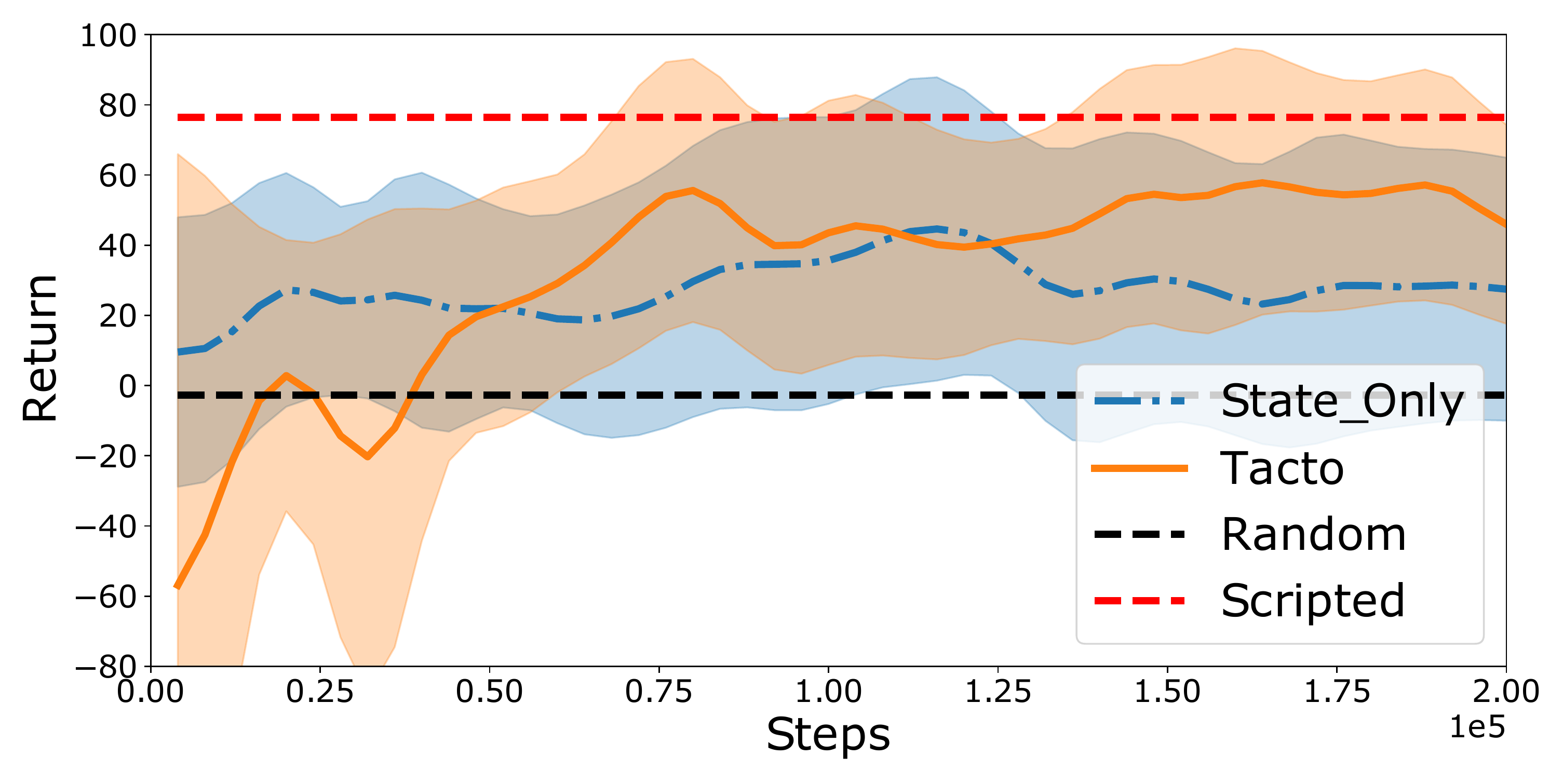}
  \caption{rhythm task}
  \label{fig:Q1_2a}
\end{subfigure}
\begin{subfigure}{0.48\columnwidth}
  \centering
  \includegraphics[width=\linewidth]{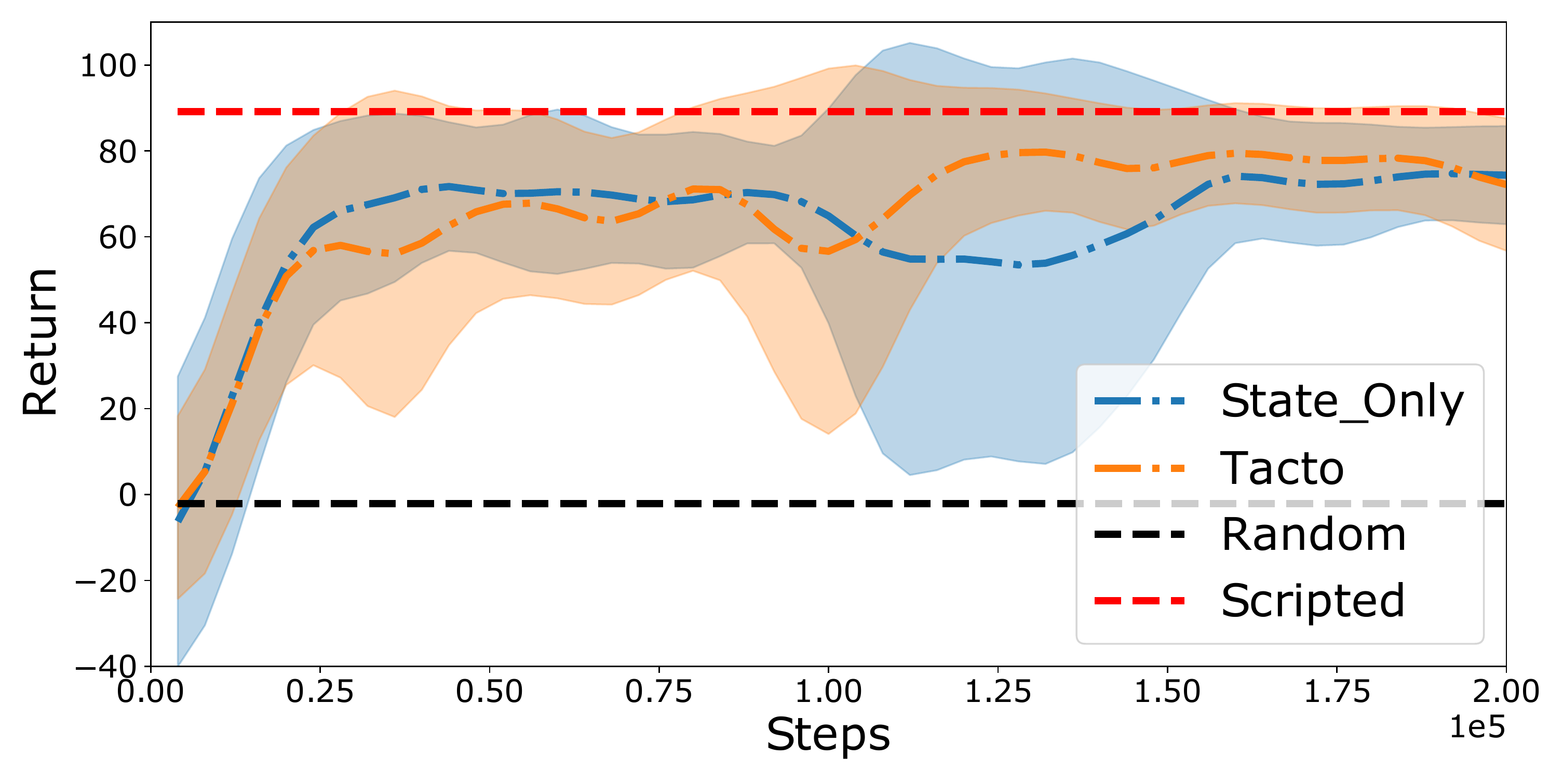}
  \caption{chord task}
  \label{fig:Q1_2b}
\end{subfigure}

\caption{Evaluation on tasks with different rhythms and chords. All the experiments are run with 5 seeds. The shade area is one standard deviation. Our algorithm can successfully accomplish harder music tasks.  }
\label{fig:Q1_2}
\end{figure}

We build an environment using the Bullet physics engine~\cite{coumans2019} in conjunction with the TACTO tactile sensors simulator~\citep{wang2020tacto}.
Our simulated environment consists of a piano keyboard and an allegro~\citep{allegro} robot hand with simulated DIGIT tactile sensors~\citep{Lambeta2020DIGIT} mounted on each fingertip. \fig{fig:demo} shows visualized samples from the environment. 
The pianos has 14 white keys and 10 black keys that have 128 discretized levels of key velocities to reflect the sound of an attacked key. We simplify the dynamics of the piano keyboard with a non-linear spring force.

The allegro hand is initialized above the keyboard with random horizontal root positions.
Each joint of the hand can be controlled; however, for most of the experiments we freeze the joints that are not crucial for the task. The hand can explore horizontally above the keyboard. We use the reward function defined in \eq{eq:gen_rew} and/or \eq{eq:w_rew} for the task. 

To evaluate the performance, we use both the accumulated reward and qualitative samples.

\subsection{Learning to Play the Piano}

We list representative tasks that mimicking realistic amateur piano players learning to play.

\noindent \textbf{One-note task:} Play one specified quarter note with corresponding velocity.

\noindent \textbf{Three-note task:} Play three specified quarter notes with corresponding velocity.

\noindent \textbf{Six-note task:} Play twelve specified quarter notes with corresponding velocity.

\noindent \textbf{Rhythmic task:} Play specified three notes that are randomly chosen from quarter notes, eighth notes and triplets.

\noindent \textbf{Chord task:} Play three chords with corresponding velocity. Two notes are played simultaneously in a chord.

We now answer the question: \textit{Can we use RL to learn to play piano?} In \fig{fig:Q1_1}, we find that the RL agent achieves reasonable returns after training 2e5 iterations for the one-note, the three-note, and the six-note tasks. The averaged returns for each note become lower when the task difficulty level increases. We attribute this to the longer horizon in harder tasks and the corresponding exploration challenge. Comparing between the RL agent with tactile image and the agent without it, we find that the tactile sensor improves the learning efficiency. However, the asymptotic performance would be similar if the agent is trained with sufficient simulation steps. We also compare the RL-based agents with random controllers and scripted agents. We find that RL-based agents outperform the random controller by a large margin while achieving similar performance as the scripted agents. We note that the scripted agents are programmed with human priors to move the hand to hit all the required keys. 

In \fig{fig:Q1_2}(a), we experiment with different rhythms that includes eighth notes~(half-beat). We find that the proposed methods have the ability not only to handle quarter notes, but also a mixture of different note durations. In \fig{fig:Q1_2}(b), we also experiment with the chord task to play multiple notes simultaneously. We find that the agent can successfully play the first two chords. For the third chord, there is occasionally one wrong note. We attribute this to the lack of exploration for the end of a task. This might be fixed by better exploration strategy and more exploration steps. 


\subsection{Piano Playing Agent with Fingering Indicator}

\begin{figure*}[t]
  \centering
  \includegraphics[width=0.98\linewidth]{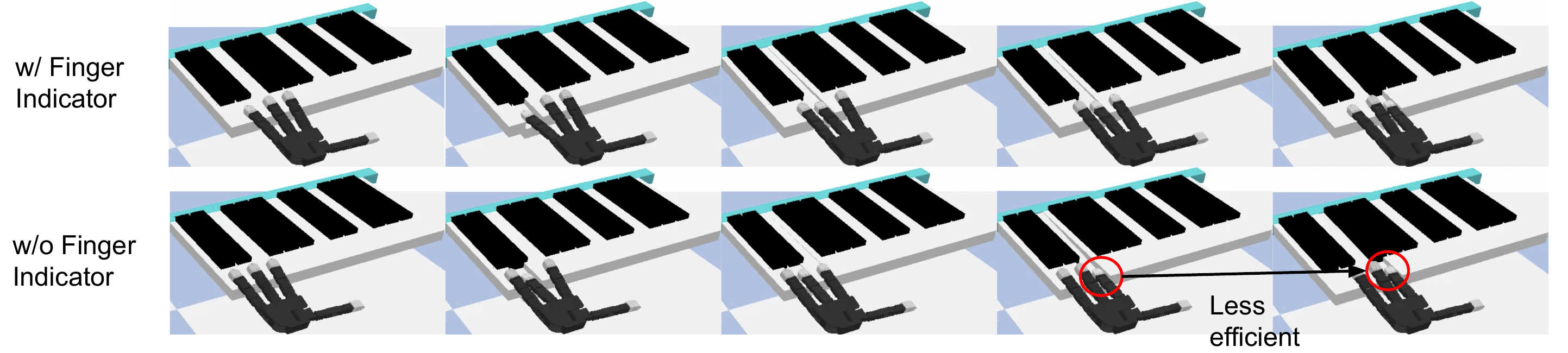}
  \caption{Qualitative results for training with or without tactile fingering indicators. In the experiments, we find that the robot hand trained without fingering indicators will easily converge to local optimum using wrong fingering. The robot hand trained with tactile fingering indicators is playing the note with the most efficient finger usage. This is similar to human piano learners who has a hard time understanding the best fingering for a specific piece. }
\label{fig:viz}
\end{figure*}

In this section, we investigate whether the agent can play the notes correctly with tactile-based fingering indicator. The fingering indicator provides information for the agents about which finger to use. The provided indicators usually optimize a long-horizon cost. For example, the correct fingering would cause less overall hand movement. 
We apply the same algorithm with the reward function described in \sec{sec:approach}. When the indicator is provided, the agent will only be rewarded if it plays the correct key with the correct finger. We note that a well-trained agent would have similar returns as there is no fingering indication reward. However, qualitatively the results can be different. In \fig{fig:viz}, we demonstrate that the proposed reward results in correct and more efficient fingering. The agent without fingering indication reward cannot find the correct fingering. This is very similar to piano learning in the real world where amateur players have hard times finding the most efficient fingering without finger indication. One potential future direction would be encouraging RL agents to explore better fingering for playing a specific piece.
\begin{table}
\centering
\caption{Evaluation of different reward functions. All the experiments are run with 5 seeds. One standard deviation is shown in the parenthesis. The results are evaluated with the Linear reward.}\label{tab:rew}
\begin{tabular}{l|ccccccccc}
\toprule
 & Avg. Return (Std)   \\
\midrule
Linear    & $35.21 (19.32)$ \\
Squared     & $17.55 (16.12)$  \\
Wasserstein     & $\mathbf{43.03  (20.08)}$  \\
\bottomrule
\end{tabular}
\end{table}

\begin{table}
\centering
\caption{Evaluation on different exploration strategies. All the experiments are run with 5 seeds. One standard deviation is in the parenthesis. Adding more priors to the exploration strategy would significantly improve the performance.}\label{tab:explore}
\begin{tabular}{l|ccccccccc}
\toprule
 & Avg. Return (Std)   \\
\midrule
Prior Explore    & $\mathbf{119.37 (61.2)}$ \\
Uniform Move Hand     & $67.82 (61.3)$  \\
No Action Repeat     & $74.52  (39.4)$  \\
\bottomrule
\end{tabular}
\end{table}
\subsection{Ablation Study}
\subsubsection{Reward Design}
We compare the reward described in~\sec{sec:reward}. Specifically, we have three reward functions: 1) linear difference reward, 2) squared difference reward, 3) wasserstein reward. 

In \tab{tab:rew}, we compare the results based on different reward functions. We note that since different reward functions cannot be compared directly, we test the trained agents on the linear reward. We find that the wasserstein reward outperforms the other two rewards even evaluated based on linear rewards. This is because wasserstein reward has dense feedback and hence can guide the agent to better policy. We also find that the squared difference loss has slightly lower performance than the linear difference reward. This might be due to the wrong scale of the reward function.



\subsubsection{Exploration Strategy}
We ablate the exploration strategies that has different amount of priors. Specifically, we have three exploration strategies: 1) uniform action where all the actions are sampled uniformly, 2) uniform action with repeats where each action has a certain probability to repeat the previous action, 3) action repeats and wrist stabilizer where we add a constraint that the agent's wrist cannot move too often based on 2). We note that these exploration strategies are with stronger human priors. However, these priors is very intuitive based on human experience and easy to implement. From \tab{tab:explore}, we can observe that human priors in the exploration strategy improves the performance by a large margin. We use these exploration strategies to collect data before training starts. We believe stronger priors might lead to even better performance; however, we only implement the aforementioned natural ones. 

\subsubsection{Controlling Method}
\begin{figure}[t]
  \centering
  \includegraphics[width=\linewidth]{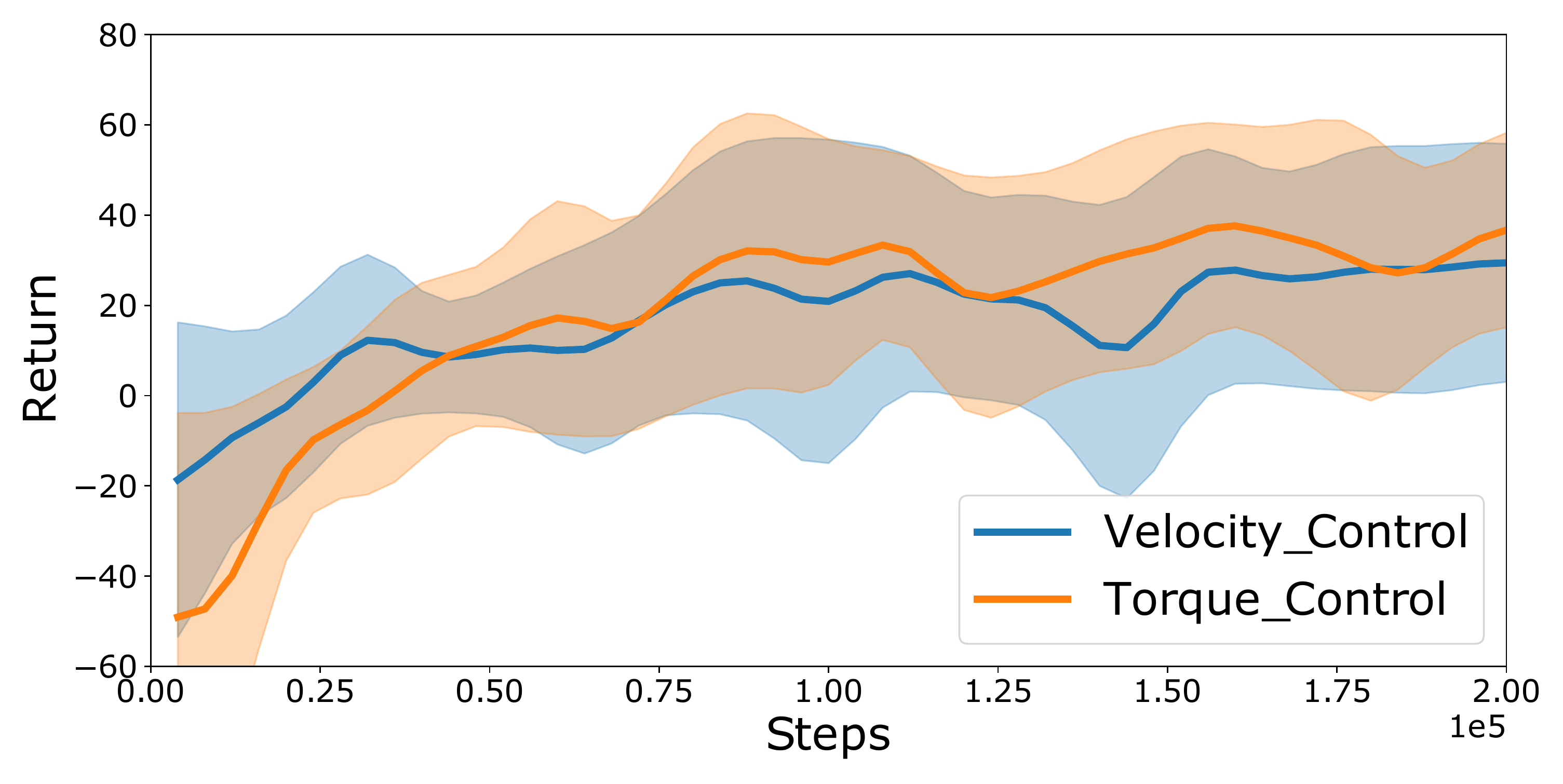}

\caption{Comparison of using torque control versus velocity control. All the experiments are run with 5 seeds and the shaded area is one standard deviation. The results show that, in simulation, the RL algorithm can learn to solve the task in either of the two modes without significant difference in performance.}
\label{fig:torque_vel}
\end{figure}

We compare the control mode for the tasks including velocity control and torque control. In \fig{fig:torque_vel}, we find that there is no significant difference between different control modes. Although it is usually harder to train RL with torque control, in the piano tasks they perform similarly. 



\begin{figure}[t]

  \centering
  \includegraphics[width=0.96\linewidth]{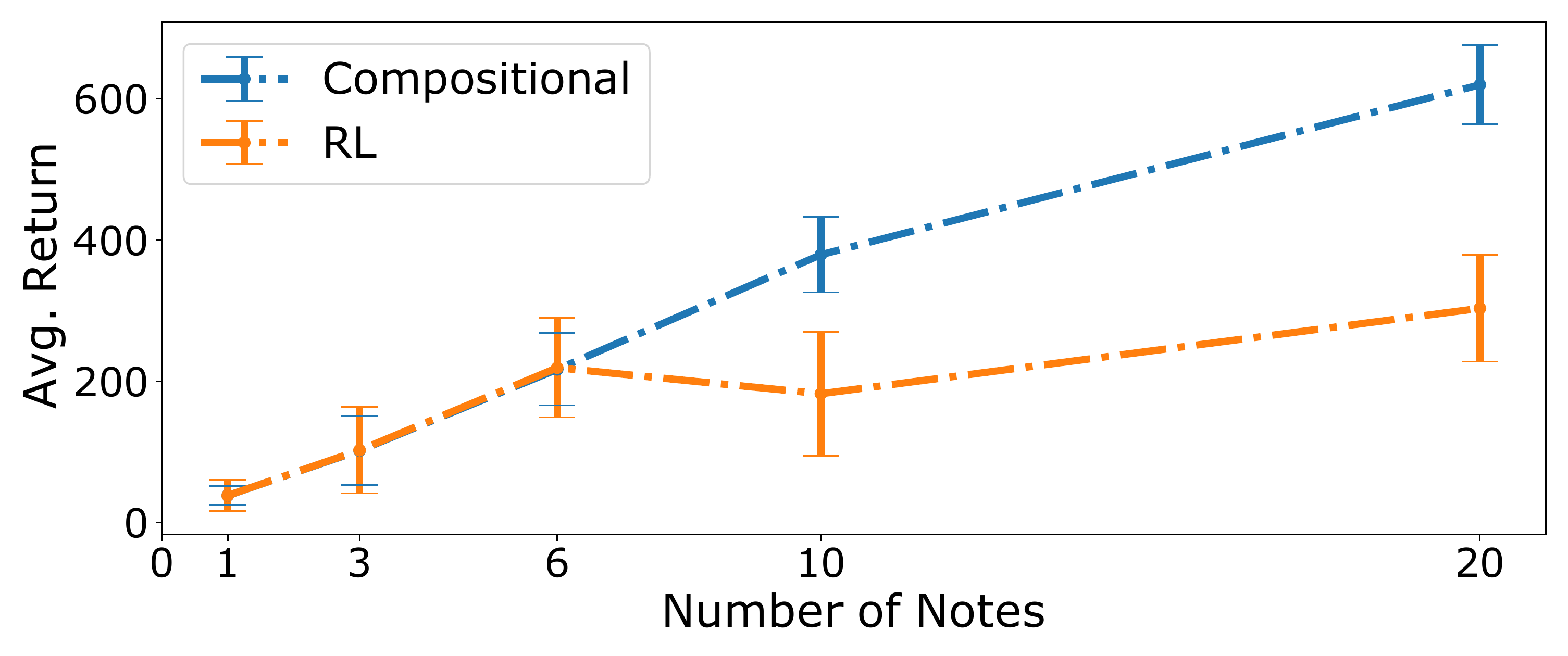}
  \caption{Comparison between compositional execution of policies and end-to-end trained policy for the task. We observe that when playing longer executions, our compositional approach outperform the simple end-to-end learning approach. }
\label{fig:comp}

\end{figure}

\subsection{Compositional Execution for long-horizon task}
Many realistic piano pieces are long in time. Hence, RL algorithms might suffer from the long horizon nature and fail to learn to play the required notes. In this section, we answer the question on how can RL agents deal with the long-horizon nature of music tasks. However, it is natural to learn an easier task within a shorter horizon $L$ and repetitively execute the policy with a sliding window only showing the composition that's within $L$. In \fig{fig:comp}, we find that when the horizons of tasks are short, there is no difference between RL and compositional execution while compositional execution is superior when the horizon is significantly longer. This is due to the hard exploration problem in long horizon tasks. This also shed some lights on future research on this track that we can train well an RL agents within a phrase or a few notes and execute the whole piece with one policy.


%


\section{CONCLUSION}
\label{sec:conclusion}

	In this paper, we proposed the first reinforcement learning-based approach for learning to play piano with robot hands equipped with tactile sensors and built the hand-piano simulator. To tackle this challenge, we designed a general reward function and a series of curriculum tasks for piano playing. The empirical results show that a simulated robot hand can play a piano piece with correct notes, velocity and fingering.








\bibliographystyle{IEEEtran}
\bibliography{paper}

\end{document}